\documentclass[graybox]{svmult}

\usepackage{type1cm}        
\usepackage{makeidx}         
\usepackage{graphicx}        
\usepackage{multicol}        
\usepackage[bottom]{footmisc}
\usepackage{newtxtext}       %
\usepackage{newtxmath}       
\usepackage{url}

\makeindex             
\usepackage{booktabs}
\usepackage{colortbl}
\usepackage{multirow}
\usepackage{multicol}
\usepackage{diagbox}
\usepackage{makecell}

\usepackage{xcolor}
\definecolor{mygreen}{rgb}{0.33,0.6,0.18}
\arrayrulecolor{mygreen}

\newcolumntype{L}[1]{>{\raggedright\let\newline\\\arraybackslash\hspace{0pt}}m{#1}}
\newcolumntype{C}[1]{>{\centering\let\newline\\\arraybackslash\hspace{0pt}}m{#1}}
\newcolumntype{R}[1]{>{\raggedleft\let\newline\\\arraybackslash\hspace{0pt}}m{#1}}

\newcommand{\ie}{\textit{i}.\textit{e}.}
\begin{document}

\title*{Automatically Extracting Information in Medical Dialogue: Expert System And Attention for Labelling}
\titlerunning{Automatically Extracting Information in Medical Dialogue ...}
\author{Xinshi Wang, Xunzhu Tang}
\institute{Xinshi Wang \at Rensselaer Polytechnic Institute, 110 8th St, Troy, NY 12180, \email{wangx47@rpi.edu}
\and Xunzhu Tang \at University of Luxembourg, 2 Av. de l'Universite, 4365 Esch-sur-Alzette, Luxembourg, \email{xunzhu.tang@uni.lu}}
%
%
\maketitle

\abstract*{Each chapter should be preceded by an abstract (no more than 200 words) that summarizes the content. The abstract will appear \textit{online} at \url{www.SpringerLink.com} and be available with unrestricted access. This allows unregistered users to read the abstract as a teaser for the complete chapter.
Please use the 'starred' version of the \texttt{abstract} command for typesetting the text of the online abstracts (cf. source file of this chapter template \texttt{abstract}) and include them with the source files of your manuscript. Use the plain \texttt{abstract} command if the abstract is also to appear in the printed version of the book.}

\abstract{Medical dialogue information extraction is becoming an increasingly significant problem in modern medical care. It is difficult to extract key information from electronic medical records (EMRs) due to their large numbers. Previously, researchers proposed attention-based models for retrieving features from EMRs, but their limitations were reflected in their inability to recognize different categories in medical dialogues. In this paper, we propose a novel model, Expert System and Attention for Labelling (ESAL).  We use mixture of experts and pre-trained BERT to retrieve the semantics of different categories, enabling the model to fuse the differences between them.  In our experiment, ESAL was applied to a public dataset and the experimental results indicated that ESAL significantly improved the performance of Medical Information Classification. The code is available at here. \footnote{\url{https://github.com/Xinshi0726/Expert-System-and-Attention-for-Labelling}}
}

\section{Introduction}
Increasingly, hospitals are prioritizing Medical Dialogue Information Extraction (MDIE) due to the adoption of Electronic Health Records (EHR). Using MDIE, detailed medical information can be extracted from doctor-patient conversations. MDIE can be viewed as a multi-label classification problem made up of different classes and their status labels. Specifically, the dataset we used in this paper includes symptoms, surgeries, tests, and other information.

Medical dialogue information extraction has received an increasing amount of attention from scholars, and various approaches have been developed. Doctor-patient dialogues were firstly converted to electronic medical records and the medical dialogue information extraction task was introduced, but no specific model to solve the task was proposed \cite{finley2018automated}. As a result, 186 symptom codes and their corresponding statuses were defined in a new dataset and proposed as a new task. By proposing two novel models, this problem was solved \cite{du2019extracting}. The first model was a span-attribute tagging model, and the second was a sequence-to-sequence model. Even though a wide range of symptoms was covered in the dataset, other critical medical information wasn't considered. To incorporate more medical information, a novel dataset that includes four main categories, namely symptoms, surgeries, tests, and other information, was introduced \cite{zhang2020mie}.
Furthermore, several specific items with corresponding statuses were predefined. In addition, a novel method of annotation was proposed, the sliding window technique, so that the dialogues included within the document could contain the proper amount of information. Meanwhile, a Medical Information Extractor (MIE) for multi-turn dialogues was developed \cite{zhang2020mie}. A matching mechanism was used to match dialogues between predefined category-item representations and status representations. The utterance's category-item information was exploited to match its most suitable status in a window to aggregate its category-item and corresponding status information.

With the help of mixture of experts \cite{jacobs1991adaptive, ma2018modeling, zhuang2020comprehensive}, we propose a model called Expert System and Attention for Labeling (ESAL) that extracts various representations of dialogue to address the different categories within the dialogue. To get category-specific representations, we first use BERT \cite{devlin2018bert} to extract contextual representations of the dialogue and feed them to the category-specific BiLSTM \cite{schuster1997bidirectional} expert. After that, we calculate the attention value between the encoded candidate representation and the encoded dialogue representation in order to obtain the candidates. In a similar manner, we calculate the status using the same attention mechanism. 

To summarize, this paper makes the following contributions:

\begin{itemize}
  \item This paper proposes an expert system attention for labelling model for extracting medical dialogue information. Each specified category can be captured in terms of the utterance representation.
  \item In this study, we introduce an expert system that effectively strengthens the understanding of doctor-patient dialogue. To facilitate understanding, we also introduce a learnable embedding layer.
  \item On a widely used medical dialogue dataset, we perform extensive experiments. On window-level evaluation, our model achieves an F1 score of 70.00, while on dialogue-level evaluation, it scores 72.17. On the benchmark dataset, it outperforms the state-of-the-art approaches by a significant difference, demonstrating its effectiveness.
\end{itemize}

\section{Related Work}

\subsection{Medical Dialogue Information Extraction}

Medical Dialogue Information Extraction has attracted increasing scholar attention due to the growing priority of building Electronic Health Records in hospitals \cite{wachter2018combat, xu2018burnout}. 
To address this problem, a dataset with 4 predefined categories: \ie, symptom, test, surgeries, and other information, as well as their corresponding status was proposed \cite{zhang2020mie}. The dataset can be viewed as a multi-label classification problem: there is a multi-label binary representation of the predefined category with its corresponding status for each doctor-patient dialogue window. The task takes a doctor-patient dialogue window as input and expects a multi-label binary representation of the category status pair as output. Each multi-label binary representation should have length equal to 355 as it is the number of elements in the Cartesian product of the items in the 4 categories with their corresponding status. To perform classification on the entire dialogue, results from each window will be merged to form a new set.

\subsection{Mixture of Experts}
Mixture of Experts is composed of many separate networks, each of which learns to handle a subset of the complete set of training cases \cite{jacobs1991adaptive}. The ensemble of individual experts has proven to be able to improve performance \cite{caruana1993multitask, hinton2015distilling}. Then, mixtures of experts system were converted into a basic building block \cite{eigen2013learning, shazeer2017outrageously}. Mixture of Experts has been applied to various fields, such as multi-domain fake news detection \cite{nan2021mdfend} and recommendation systems \cite{ma2018modeling}.

\begin{figure*}[ht]
\centering
\includegraphics[width=\linewidth, scale = 0.5]{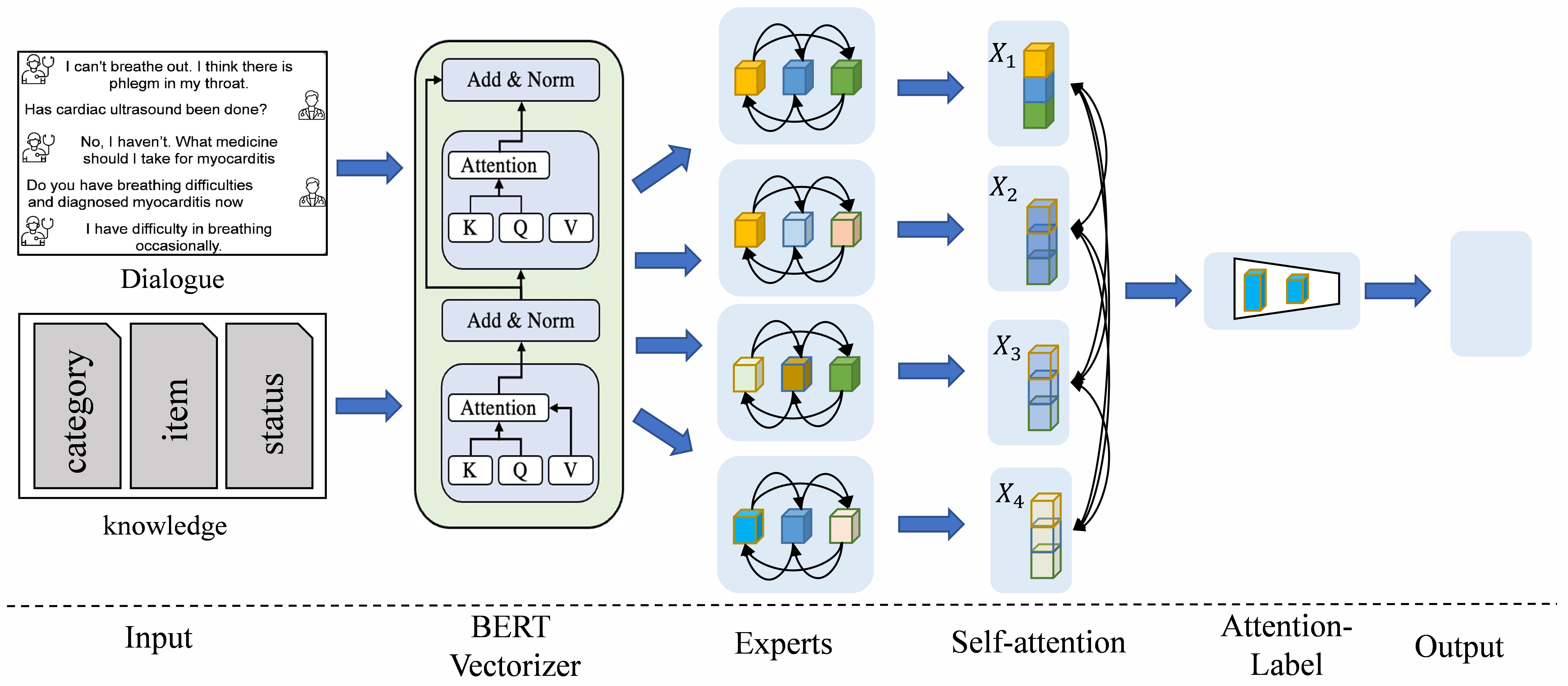}
\caption{Model Architecture}
\label{fig:architecture}
\end{figure*}

\section{Approach}
In this section, we will elaborate the architecture of ESAL.
The architecture is shown in figure \ref{fig:architecture}. ESAL is composed of 4 different stages: 1). Embedding layer 2). Expert information extraction Layer 3). Self-Attention labelling Layer  4). Output Layer.

\subsection{Embedding layer}
For each doctor-patient dialogue, we first tokenize its content with Bert Tokenizer \cite{devlin2018bert}. We then add special tokens for classification (\ie,$[cls]$) as well as separation (\ie,$[sep]$) to obtain a list of tokens $X = [[cls],token_1, token_2,...,token_n,[sep]]$. We then feed the list of tokens into BERT to obtain word embedding $V = BERT(X)$. Similarly, we perform the same operation on the candidates for matching to obtain the embedding $U = BERT(Q)$ for query $Q$.

\subsection{Expert information extraction layer}
With the advantage of Mixture-of-Experts, we employ multiple experts (i.e., network) to extract category-specific and status-specific representations of the utterance. We select the bidirectional long short-term memory network (BiLSTM) \cite{schuster1997bidirectional} with attention mechanism \cite{vaswani2017attention} as our individual network. BiLSTM has been widely used to extract contextual text features. 

The equation below denotes the process for encoding each dialogue, where $H_{C}[i]$ consists of the contextual representation of embedding $V$ specific to category $i$ and $H_{S}$ consists the status representation of embedding $V$.
\begin{equation} \label{eq1}
\begin{split}
H_{C}[i],H_{S} & = BiLSTM(V), BiLSTM(U)
\end{split}
\end{equation}

For candidates in the form of $\{Category:Item-Status\}$, we denote the Cartesian product between item and status given the category as $Q_{C}$. 
We then feed $Q_C$ to the corresponding category expert to obtain the embedding and apply self-attention to the embedding to obtain a single vector $C_{C}$ that compresses the information of the entire sequence in a weighted way. The procedures above can be described with the following equation, where $\sigma = \frac{exp(i)}{\sum_{i = 1}^{n}\exp(i)}$ denotes the softmax operation.

\begin{equation} \label{eq1}
\begin{split}
U_{C}[i],U_{S} & = BiLSTM(Q_{C}), BiLSTM(Q_{S})\\
A_{C}[i],A_{S} & = WU_{C}[i]+b, WU_{S}+b\\
P_{C}[i],P_{S} &= \sigma(A_{C}[i]), \sigma(A_{S})\\
C_{C},C_{S} &= \sum_{i}^{n}(P_{C}[i]U_{C}[i]), \sum_{i}^{n}(P_{S}U_{S})\\
\end{split}
\end{equation}

\subsubsection{Domain Gate}
To incorporate information from all domains, we propose a domain gate with category representations from all domains as input. The output of the domain gate is the vector $H_C$ indicating the weight ratio of each expert. Let $Gate(\cdot)$ denote the gate operation, we can describe the domain gate as the following equation:

\begin{equation} 
\begin{split}
H_C &= Gate(\sum_{i=1}^4 H_C[i])
\end{split}
\end{equation}
where the $Gate(\cdot)$ operation is a feed-foward network.

\subsection{Self-Attention Labeling Layer}
We employ self-attention to capture the most relevant candidate features from the utterance representation, where the candidate representation is treated as a query to calculate the attention value $Q_{C}$ towards the category-specific utterance representation. Similarly, the candidate status representation is treated as another query to calculate the attention value toward the original utterances to obtain the most relevant status features from utterance representation. The process can be described with following equation: 

\begin{equation}
\begin{split}
    P_{C}[i],P_{S}[i] &= \sigma(C_{C}[i]H_{C}), \sigma(C_{S}[i]H_{S})\\
    Q_{C}[i],Q_{S}[i] &= \sum_{j}^{n}(P_{C}[i,j]H_{C}[j]), \sum_{j}^{n}(P_{S}[i,j]H_{S}[j])\\
\end{split}
\end{equation}

To assign the correct candidates to each dialogue window, we need to match every $Q_{C}[i]$ with the corresponding $Q_{S}[i]$.  The category-item pair information and the status information does not necessarily appear in the same dialogue window, so we need to take the interactions between utterances among multiple dialogue windows into consideration. The process can be described with following equation, where $concat$ denotes the concatenate operation:

\begin{equation}
\begin{split}
   C[i] &= \sigma(Q_C[i] WQ_S[i])\\
   \hat{Q}_S[i] &= \sum_{j = 1}^{n}(C[i,j]Q_S[i])\\
   F[i] &= concat(C[i],\hat{Q}_S[i])
\end{split}
\end{equation}

The output of the equation above gives the candidate information assigned to the doctor patient dialogue $U$.

\subsection{Output Layer}
We use the output from the Self-Attention Labeling Layer, (\ie F[i]) to generate the output for our model. Using a feedforward network, we can project the utterance's representation F[i] onto the 355 corresponding candidate positions, and then apply a softmax function to select the final prediction label. The process can be described with the following equations, where $f$ denotes the feedforward network and $h_ \theta (x) =  \frac{\mathrm{1} }{\mathrm{1} + e^- \theta^Tx }$ denotes the sigmoid function: 
\begin{equation}
\begin{split}
   s[i] &= f(F[i])\\
   y &=  h_\theta(max(s[i]))\\
\end{split}
\end{equation}

\subsection{Loss Function}

We adopt the cross entropy loss as our loss function. The function is defined as the following equation: 

\begin{equation}
\begin{split}
   L = \frac{1}{I \times J} \sum_{i} \sum_{j} -y_{j}^{i} \ln{(\hat{y_{j}^{i}})} + (1-y_{j}^{i})\ln{(1-\hat{y_{j}^{i}})}
\end{split}
\end{equation}

The $y_{j}^{i}$ is a binary encoding that denotes label of $j^{th}$ candidate from the $i^{th}$ label.I denotes the number of samples and J denotes the number of candidates. $\hat{y_{j}^{i}}$ denotes the ground truth value of label $y_{j}^{i}$.

\section{Experiments}
In this section, we will conduct experiments on the MIE dataset \cite{zhang2020mie}. We will firstly describe the dataset and evaluation metrics. Then we will present results with a case study of the experiment.

\subsection{Dataset Description}
We evaluate our model on a public dataset MIE \cite{zhang2020mie}. An example of a dialogue window is illustrated in Table \ref{table:dialogue} below. 

\begin{table}[ht]
    \caption{Dialogue Window}
    \centering
    \begin{tabular}{l|r}
    \toprule
    Role & Dialogue \\ \hline
    \midrule
    {\bf Patient:} & Doctor, is it premature beat?\\
    {\bf Doctor:} & Yes, Do you feel short breath?\\
    {\bf Patient:} & No. Should I do radio frequency ablation?\\
    {\bf Doctor:} & You should. Any discomfort in chest?\\
    {\bf Patient:} & I always have bouts of pain.\\
    \bottomrule
    \end{tabular}
    \label{table:dialogue}
\end{table}

The annotation of the sliding window dialogue is composed of several labels in the form of $\{Category:Item-Status\}$. An example of the annotated label is given in table \ref{table:annotation}. 

$Category$ contains four main categories (Symptom, Surgery, Test, and Other Info). $Item$ stands for the frequent items with respect to each category. There are 45, 4, 16, and 6 items, respectively. The $status$ is defined as doctor-pos, doctor-neg, patient-pos, patient-neg, or unknown. There are in total 1,120 dialogues, resulting in 18,212 windows. The data is divided into train/develop/test sets of size 800/160/160 for dialogues and 12,931/2,587/2,694 for windows respectively. In total, there are 46,151 annotated labels, averaging 2.53 labels in each window, 41.21 labels in each dialogue.

\begin{table}[ht]
    \caption{Dialogue Annotation}
    \centering
    \begin{tabular}{l|r}
    \toprule
    Category & Item(Status) \\ \hline
    \midrule
    {\bf Symptom:} & high blood pressure (doctor-pos) \\
    {\bf Symptom:} & heart disease (unknown)\\
    {\bf Test:} & electrocardiogram (pos)\\
    \bottomrule
    \end{tabular}
    \label{table:annotation}
\end{table}
\subsection{Evaluation Metrics}
We use the precision, recall, and F1 score to evaluate our results. We also follow the evaluation metrics MIE \cite{zhang2020mie} employed to further analyze the model behavior from easy to hard. Category performance considers the correctness of the category. Item performance considers the correctness of the category and the item. Finally, the Full category takes the category, item, and the status into consideration, meaning all of them have to be completely correct. We will report the results in both the window-level and the dialogue level to further examine our results.

Window-level: The results of each segmented window are evaluated and reported by the micro-average of all windows in the test set. Category evaluation means a prediction is assumed correct if the category matches the ground truth value. Item means a prediction is assume correct if both the category and the item match the ground truth value. Full evaluation is assumed correct if the category, item, and status match the ground truth value at the same time.

Dialogue-level: We merge the results with the same category and item of all the windows in the same dialogue. For category-item pair with multiple status assigned, we replace the unknown status with any other status occurred and replace the negative status with positive status if occurred. 

\begin{figure}[ht]
\centering
\includegraphics[scale = 0.1]{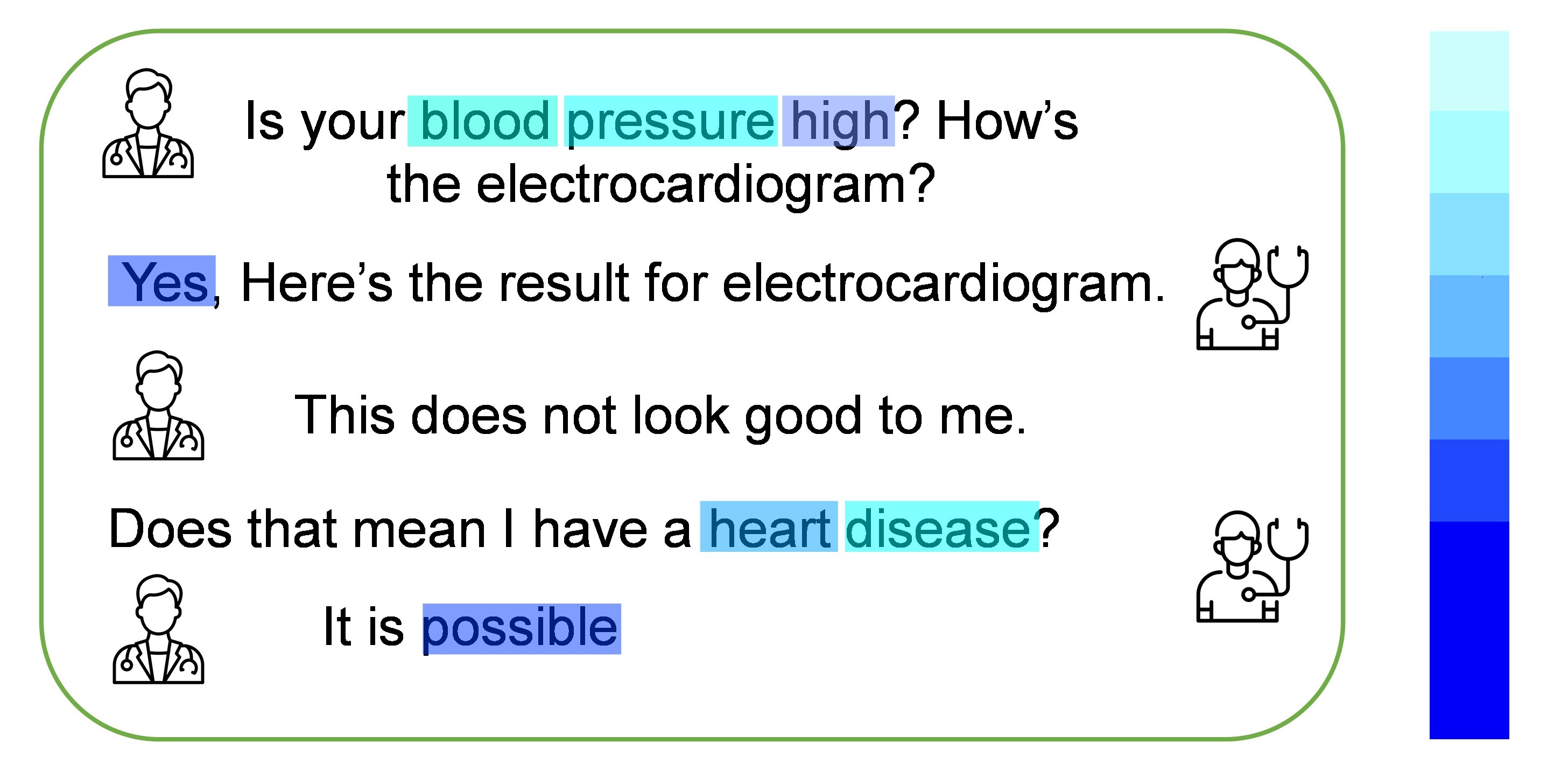}
\caption{Symptom Expert Attention Heat Map}
\label{fig:sym}
\end{figure}

\begin{figure}[ht]
\centering
\includegraphics[scale = 0.1]{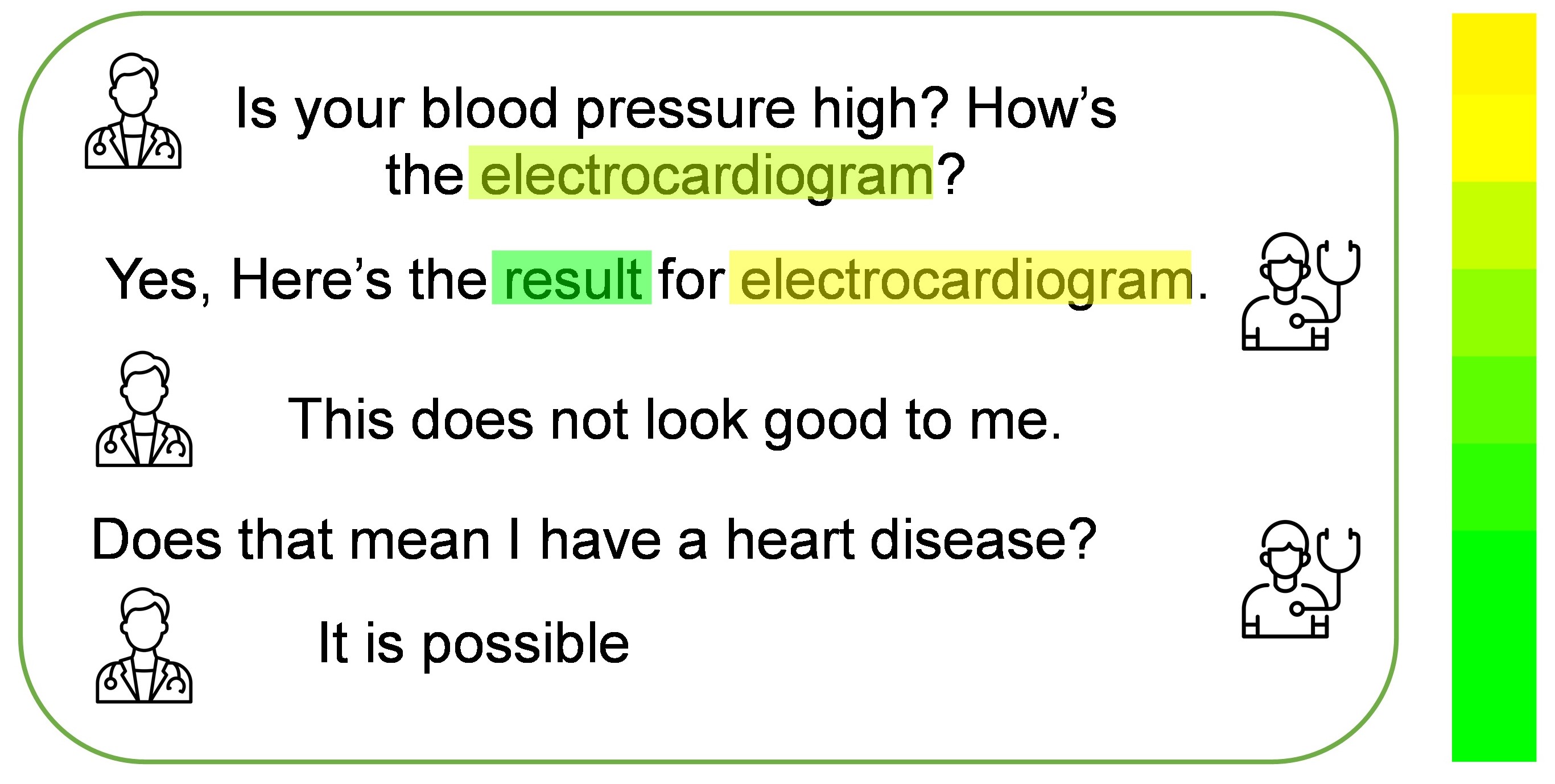}
\caption{Test Exper Attention Heat Map}
\label{fig:test}
\end{figure}
\subsection{Main Results}

\begin{table*}[ht]
\centering
\resizebox{\textwidth}{!}{
\begin{tabular}{l|lll|lll|lll}
\label{table:result}
 \diagbox{{\bf Models}}{{\bf Levels}}&
  \multicolumn{9}{c}{Window-Level} \\ \hline
\multirow{2}{*}{Model} &
  \multicolumn{3}{c}{Category} &
  \multicolumn{3}{c}{Item} &
  \multicolumn{3}{c}{Full} \\
 &
  \multicolumn{1}{c}{P} &
  \multicolumn{1}{c}{R} &
  \multicolumn{1}{c}{F1} &
  \multicolumn{1}{c}{P} &
  \multicolumn{1}{c}{R} &
  \multicolumn{1}{c}{F1} &
  \multicolumn{1}{c}{P} &
  \multicolumn{1}{c}{R} &
  \multicolumn{1}{c}{F1} \\
Plain-Classifier &
  67.21 &
  63.78 &
  64.92 &
  60.89 &
  49.20 &
  53.81 &
  53.13 &
  49.46 &
  50.69 \\
MIE-Classifier-Single &
  80.51 &
  76.39 &
  77.53 &
  76.58 &
  64.63 &
  68.30 &
  68.20 &
  61.60 &
  62.87 \\
MIE-Classifier-Multi &
  80.72 &
  77.76 &
  78.33 &
  76.84 &
  68.07 &
  70.35 &
  67.87 &
  64.71 &
  64.57\\
MIE-Single &
  78.62 &
  73.55 &
  74.92 &
  76.67 &
  65.51 &
  68.88 &
  69.40 &
  64.47 &
  65.18 \\
MIE-Multi &
  80.42 &
  76.23 &
  77.77 &
  77.21 &
  66.04 &
  69.75 &
  70.24 &
  64.96 &
  66.40 \\
ESAL &
  \bf{92.42} &
  \bf{89.66} &
  \bf{90.26} &
  \bf{89.46} &
  \bf{83.38} &
  \bf{84.85} &
  \bf{72.08} &
  \bf{70.93} &
  \bf{70.00} \\ \hline
\end{tabular}
}
\caption{Window-Level Evaluation Result, the results for MIE models are adopted from \cite{zhang2020mie}}
\label{table:result}
\end{table*}

\begin{table*}[ht]
\centering
\resizebox{\textwidth}{!}{
\begin{tabular}{l|lll|lll|lll}
\label{table:result}
 \diagbox{{\bf Models}}{{\bf Levels}}&
  \multicolumn{9}{c}{Dialogue-Level} \\ \hline
\multirow{2}{*}{Model} &
  \multicolumn{3}{c}{Category} &
  \multicolumn{3}{c}{Item} &
  \multicolumn{3}{c}{Full} \\
 &
  \multicolumn{1}{c}{P} &
  \multicolumn{1}{c}{R} &
  \multicolumn{1}{c}{F1} &
  \multicolumn{1}{c}{P} &
  \multicolumn{1}{c}{R} &
  \multicolumn{1}{c}{F1} &
  \multicolumn{1}{c}{P} &
  \multicolumn{1}{c}{R} &
  \multicolumn{1}{c}{F1} \\
Plain-Classifier &
  93.57 &
  89.49 &
  90.96 &
  83.42 &
  73.76 &
  77.29 &
  61.34 &
  52.65 &
  56.08 \\
MIE-Classifier-Single &
  97.14 &
  91.82 &
  93.23 &
  91.77 &
  75.36 &
  80.96 &
  71.86 &
  56.67 &
  61.78 \\
MIE-Classifier-Multi &
  96.61 &
  92.86 &
  93.45 &
  90.68 &
  82.41 &
  84.65 &
  68.86 &
  62.50 &
  63.99 \\
MIE-Single &
  96.93 &
  90.16 &
  92.01 &
  94.27 &
  79.81 &
  84.72 &
  75.37 &
  63.17 &
  67.27 \\
MIE-Multi &
  \bf{98.86} &
  91.52 &
  92.69 &
  \bf{95.31} &
  82.53 &
  86.83 &
  \bf{76.83} &
  64.07 &
  69.28 \\
ESAL &
  96.51 &
  \bf{95.05} &
  \bf{94.74} &
  92.52 &
  \bf{90.88} &
  \bf{90.50} &
  73.68 &
  \bf{73.10} &
  \bf{72.17} \\ \hline
\end{tabular}
}
\caption{Dialogue-Level Evaluation Result, the results for MIE models are adopted from \cite{zhang2020mie}}
\label{table:result}
\end{table*}
The experimental results are show in Table \ref{fig:architecture}. From the table, we can make the following observations.

     On both the window-level and dialogue level evaluation, Our model outperforms other models in most metrics. On window-level Full evaluation, our method has the performance improved by $5.4\%$ compared to the MIE-multi in F1 score. On dialogue-level full evaluation, our method achieves an improvement of $4.17\%$ in F1 score. These results demonstrate that the ESAL is performing better compared to the previous state-of-the-art model.
    
    On Window-level evaluation, our model outperforms other models significantly in Category and Item evaluation. For Category evaluation, Our model has a performance improvement of $16.90\%$ in F1 score. For Item evaluation, our model has a improvement of $21.65\%$ in F1 score. Also, the improvement on Precision and Recall are significant. These results demonstrate that ESAL is able to extract a better domain-specific representation of the utterance.

\subsection{Case Analysis}
In this section, we perform an analysis on a specific case to verify the effectiveness of the mixture of experts. We did a data visualization on the attention value from Symptom expert and Test expert on the same utterance in graph \ref{fig:sym} and graph \ref{fig:test}. Brighter Color suggests a higher attention value. The label for the utterance is $\{$Symptom: high blood pressure- doctor-pos, Symptom: heart disease-unkown, Test: electrocardiogram-pos$\}$. As we can see from graph \ref{fig:sym}, the highest attention value comes from "Yes", which suggests that our Symptom Expert captures the status information correctly. It also captures the status information for heart disease. Similarly, the test expert has captured the item and status. These two outputs gave category specific attention value on different items, thus proved the effectiveness of our model in capturing category-specific representations.

\subsection{Conclusion}
In this paper, we proposes an expert system attention for labelling model for extracting medical dialogue information, which utilizes two techniques: mixture of experts and an embedding layer. Experimental results on a public available dataset have shown that ESAL has the ability to capture category specific utterance representations and has better understanding of doctor-patient dialogue compared to previous models. For future work, We plan to investigate the interaction between doctor and patient to handle the pronoun ambiguity.

\end{document}